\documentclass{bmvc2k}

\newcommand{\bg}[1]{\boldsymbol{#1}} 
\newcommand{\bm}[1]{\mathbf{#1}} 

\newcommand\T{{\mathpalette\raiseT\intercal}}
\newcommand\raiseT[2]{%
\setbox0\hbox{$#1{#2}$}\raise\dp0\box0}

\usepackage{amsmath,amssymb}
\usepackage{booktabs}

\usepackage{pifont}
\newcommand{\cmark}{\ding{51}}%
\newcommand{\xmark}{\ding{55}}%

\title{Spatio-Temporal MLP-Graph Network for 3D Human Pose Estimation}

\addauthor{Tanvir Hassan}{md.hassan@mail.concordia.ca}{1}
\addauthor{A. Ben Hamza}{hamza@ciise.concordia.ca}{1}

\addinstitution{
 Concordia University \\
 Montreal, QC, Canada
}

\runninghead{Hassan, Hamza}{MLP-Graph Network for 3D Human Pose Estimation}

\def\etal{\emph{et al}\bmvaOneDot}

\begin{document}

\maketitle

\begin{abstract}
Graph convolutional networks and their variants have shown significant promise in 3D human pose estimation. Despite their success, most of these methods only consider spatial correlations between body joints and do not take into account temporal correlations, thereby limiting their ability to capture relationships in the presence of occlusions and inherent ambiguity. To address this potential weakness, we propose a spatio-temporal network architecture composed of a joint-mixing multi-layer perceptron block that facilitates communication among different joints and a graph weighted Jacobi network block that enables communication among various feature channels. The major novelty of our approach lies in a new weighted Jacobi feature propagation rule obtained through graph filtering with implicit fairing. We leverage temporal information from the 2D pose sequences, and integrate weight modulation into the model to enable untangling of the feature transformations of distinct nodes. We also employ adjacency modulation with the aim of learning meaningful correlations beyond defined linkages between body joints by altering the graph topology through a learnable modulation matrix. Extensive experiments on two benchmark datasets demonstrate the effectiveness of our model, outperforming recent state-of-the-art methods for 3D human pose estimation.
\end{abstract}


\section{Introduction}
3D human pose estimation is a fundamental task in computer vision, with the aim of predicting the 3D pose of a human body from images or videos~\cite{martinez2017simple}. While significant strides have been made in recent years~\cite{Liu2022Survey}, accurately estimating the 3D human pose remains a challenging problem. This is largely attributed to the complex and articulated nature of the human body, as well as the difficulty of estimating 3D information from 2D images~\cite{Yuille:14,pavlakos2017coarse}.

Graph convolutional networks (GCNs) have recently emerged as a powerful framework for 3D human pose estimation~\cite{zhao2019semantic}. Despite their promising results, GCN-based methods have several limitations.  First, they use the same transformation matrix for all nodes in graph convolution, thereby limiting information exchange. To address this limitation, Liu \etal~\cite{liu2020comprehensive} introduce various weight unsharing mechanisms. One drawback of these mechanisms is that they result in a larger model size that scales with the number of body joints. Second, GCNs suffer from the oversmoothing problem~\cite{Li:18}, where the model may struggle to accurately distinguish between nodes and learn meaningful representations due to repeated graph convolutions as the network depth increases. Third, to leverage temporal correlations, these methods require significant computational resources to process a larger number of input sequences such as a 243-frame sequence. Furthermore, GCNs may not be able to capture the global contextual information or long-range dependencies between nodes in the graph, which can limit their ability to learn more complex relationships and patterns in the data.

On the other hand, Transformer architectures, which utilize a multi-head self-attention mechanism to capture both spatial and temporal correlations from sequences of 2D poses~\cite{PoseFormer:2021}, have proven effective at capturing long-range dependencies between body joints in the spatio-temporal domain. However, the complexity of the self-attention block increases quadratically with the number of input sequences, making the training and inference more computationally expensive. Taking this into account, Tolstikhin \etal~\cite{tolstikhin2021mlp} propose MLP-Mixer, which has shown competitive performance compared to more complex architectures such as Transformer networks. Compared to multi-layer perceptrons (MLPs), the MLP-Mixer model has been shown to be effective at modeling long-range dependencies in the input data. However, MLP-based models do not adequately capture the local information due largely to the lack of prior knowledge about the human skeleton topology.

In this paper, we address the aforementioned challenges by proposing a novel spatio-temporal graph neural network architecture, dubbed MLP-GraphWJ mixer, which leverages spatio-temporal correlations and also makes use of weight and adjacency modulation. The proposed framework employs a weighted Jacobi (WJ) feature propagation rule obtained via graph filtering with implicit fairing. In summary, we make the following key contributions:

\begin{itemize}
\item We propose a graph weighted Jacobi (GraphWJ) network, which employs a weighted Jacobi (WJ) feature propagation rule obtained via graph filtering with implicit fairing, and also leverages weight and adjacency modulation. 
\item We design a spatio-temporal network architecture by incorporating MLPs to capture global information and a GraphWJ network to capture local information between adjacent joints across different channels.
\item We demonstrate through experiments and ablation studies that our proposed model outperforms strong baselines, attaining state-of-the-art performance in 3D human pose estimation, while retaining a small model size.
\end{itemize}

\section{Related work}
\noindent\textbf{3D Human Pose Estimation.}\quad The basic goal of 3D human pose estimation is to estimate the 3D coordinates of the joints in the human body from images or videos. Single-stage and two-stage methods are two commonly used approaches for 3D human pose estimation. Single-stage methods are based on a direct regression from the input image to the 3D pose estimation~\cite{li20143d}. These methods typically use convolutional neural networks to extract features from the input image and then use a regression network to directly estimate the 3D pose. Two-stage methods, on the other hand, generally consist of two separate networks: an off-the-shelf 2D pose detection network to extract 2D keypoints and a 3D pose estimation network~\cite{yang20183d,fang2018learning,hossain2018exploiting,pavlakos2018ordinal,sharma2019monocular,ge20193d,pavllo20193d,zhao2019semantic,
YujunCai:19,HaiCi:2019,Kenkun:2020,zou2020high}. Two-stage methods usually achieve higher accuracy than single-stage methods, especially for complex pose estimation tasks.

\medskip\noindent\textbf{Spatio-Temporal Methods.}\quad Current monocular 3D pose estimation methods can be classified into two mainstream types: single-frame or image-based and multi-frame or video-based approaches. Single-frame-based methods aim to predict 3D pose from a single RGB image. In contrast, video-based methods take advantage of the temporal dependencies between frames in the video clip. Due to the ill-posed characteristic of generating accurate 3D poses from a single RGB image, a number of techniques~\cite{hossain2018exploiting, liu2021graph,PoseFormer:2021, Zeng2020SRNet, zeng2021learning, YujunCai:19} have been developed that rely on temporal correlations to improve the robustness and accuracy of the resulting 3D poses. Liu \etal~\cite{liu2021graph} develop graph attention blocks in conjunction with dilated temporal convolution that is capable of estimating 3D pose from consecutive 2D pose sequences. Zheng \etal~\cite{PoseFormer:2021} utilize a Transformer-based approach that is designed to capture both the correlations between human joints and their temporal dependencies. Zeng \etal~\cite{zeng2021learning} introduce a temporal aware dynamic graph convolution where the graph updates by physical skeleton topology and the features of nodes. However, most of these methods tend to be computationally demanding, and are inherently restricted in their ability to establish temporal connectivity. Moreover, most GCN-based approaches are constrained by the fact that they share a feature transformation for capturing the relationships between each node and its adjacent nodes in a graph convolution layer. Also, sharing the same feature transformation for all nodes hinders the efficient exchange of information between the body joints. Our proposed framework falls under the category of 2D-to-3D pose lifting. It differs from existing GCN-based approaches in that we use a new graph propagation rule combined with weight and adjacency modulation to learn additional connections between body joints by adjusting the graph topology through a learnable modulation matrix. We also design a network architecture that combines the strengths of MLPs and graph neural networks in order to improve accuracy in 3D human pose estimation, while maintaining simplicity and efficiency.

\section{Proposed Method}

\subsection{Preliminaries and Problem Formulation}
\noindent\textbf{Basic Notions.}\quad Let $\mathcal{G}=(\mathcal{V},\mathcal{E}, \bm{X})$ be an attributed graph, where $\mathcal{V}=\{1,\ldots,N\}$ is a set of nodes that correspond to body joints, $\mathcal{E}$ is the set of edges representing connections between two neighboring body joints, and $\bm{X}=(\bm{x}_{1},...,\bm{x}_{N})^{\T}$ is an $N\times F$ feature matrix of node attributes whose $i$-th row $\bm{x}_{i}$ is an $F$-dimensional feature vector associated to node $i$. We denote by $\bm{A}$ an $N\times N$ adjacency matrix whose $(i,j)$-th entry is equal to 1 if there the edge between neighboring nodes $i$ and $j$, and 0 otherwise. We also denote by $\hat{\bm{A}}=\bm{D}^{-1/2}\bm{A}\bm{D}^{-1/2}$ the normalized adjacency matrix, where $\bm{D}=\mathsf{diag}(\bm{A}\bm{1})$ is the diagonal degree matrix.

\medskip\noindent\textbf{Weighted Jacobi Method.}\quad Given a matrix $\bm{B}\in\mathbb{R}^{N\times N}$ and a vector $\bm{x}\in\mathbb{R}^{N}$, the weighted Jacobi iteration~\cite{Saad:03} for solving a matrix equation $\bm{B}\bm{h}=\bm{x}$ is given by
\begin{equation}
\bm{h}^{(k+1)} = \omega\,\mathsf{diag}(\bm{B})^{-1}\bm{x}+ (\bm{I}-\omega\,\mathsf{diag}(\bm{B})^{-1}\bm{B})\bm{h}^{(k)},
\end{equation}
where $\omega$ is a relaxation factor, and $\bm{h}^{(k)}$ and $\bm{h}^{(k+1)}$ are the $k$-th and $(k+1)$-th iterations of the unknown $\bm{h}$, respectively.

\medskip\noindent\textbf{Problem Statement.}\quad Let $\mathcal{D}=\left\{\left(\mathbf{x}_{i}, \mathbf{y}_{i}\right)\right\}_{i=1}^{N}$ be a training set consisting of 2D joint positions $\bm{x}_{i}\in\mathcal{X}\subset\mathbb{R}^2$ and their associated ground-truth 3D joint positions $\bm{y}_{i}\in\mathcal{Y}\subset\mathbb{R}^3$. The aim is to learn a regression model $f: \mathcal{X} \rightarrow \mathcal{Y}$ by finding a minimizer of the following loss function
\begin{equation}	
\bm{w}^{*}=\arg\min_{\bm{w}}\frac{1}{N}\sum_{i=1}^{N}l(f(\bm{x}_{i}),\bm{y}_{i}),
\end{equation}
where $l(f(\bm{x}_{i}),\bm{y}_{i})$ is an empirical loss function defined by the learning task. Since human pose estimation is a regression task, we define $l(f(\bm{x}_{i}),\bm{y}_{i})$ as a weighted sum (convex combination) of the $\ell_2$ and $\ell_1$ loss functions
\begin{equation}	
l=(1-\lambda)\sum_{i=1}^{N}\Vert\bm{y}_{i}-f(\bm{x}_{i})\Vert_{2}^{2}+ \lambda\sum_{i=1}^{N}\Vert\bm{y}_{i}-f(\bm{x}_{i})\Vert_{1},
\label{eq:loss}
\end{equation}
where $\lambda\in [0,1]$ is a weighting factor controlling the contribution of each term.

\subsection{Graph Filtering with Implicit Fairing}
In the context of graph filtering, the implicit fairing approach~\cite{Desbrun:99} is applied by defining a Laplacian operator on the graph, which captures the connectivity and structure of the graph. Specifically, graph filtering with implicit fairing can be performed by solving the sparse linear system $(\bm{I}+s\bm{L})\bm{H}=\bm{X}$, where $\bm{X}$ is the feature matrix of node attributes, $\bm{L}=\bm{I}-\hat{\bm{A}}$ is the normalized Laplacian matrix, $\bm{H}$ is the filtered graph signal, and $s$ is a positive scalar. This sparse linear system can be efficiently solved using the weighted Jacobi method~\cite{Saad:03}, which uses a weighting parameter $\omega$ to compute the $k$-th iteration as follows:
\begin{equation}
\begin{split}
\bm{H}^{(k+1)} & = \omega (\mathsf{diag}(\bm{I}+s\bm{L}))^{-1}\bm{X} + \bigl(\bm{I}-\omega(\mathsf{diag}(\bm{I}+s\bm{L}))^{-1}(\bm{I}+s\bm{L})\bigr)\bm{H}^{(k)}\\
& = \bm{H}^{(k)}-\omega\bm{H}^{(k)} + (1-\alpha)\omega\hat{\bm{A}}\bm{H}^{(k)} + \alpha\omega\bm{X}
\end{split}
\label{Eq:WJ_iterate}
\end{equation}	
where $\alpha = 1/(1+s)$, and $\omega$ can be chosen to optimize the convergence speed of the method.

\subsection{Graph Weighted Jacobi Network}
In graph neural networks, the aim of a layer-wise propagation rule is to encourage the network to learn useful representations at each layer that can be used to inform subsequent layers and ultimately improve the accuracy of the network's predictions. Drawing inspiration from the weighted Jacobi iterative solution for graph filtering with implicit fairing, we define a weighted Jacobi ($\mathsf{WJ}$) layer-wise propagation rule as
\begin{equation}
\bm{H}^{(\ell+1)} = \sigma(\mathsf{WJ}\bigl(\bm{H}^{(\ell)})\bigr),\quad \ell=0,\dots,L-1
\label{Eq:PropRule}
\end{equation}
where $\sigma(\cdot)$ is an activation function such as the Gaussian Error Linear Unit (GELU)~\cite{hendrycks2016gaussian} and $L$ is the number of layers. The input of the first layer is the initial feature matrix $\bm{H}^{(0)}=\bm{X}$. The weighted Jacobi operation on the input feature matrix $\bm{H}^{(\ell)}$ of the $\ell$-th layer is given by
\begin{equation}
\begin{aligned}
\mathsf{WJ}(\bm{H}^{(\ell)}) & = \bm{H}^{(\ell)}\bm{W}_1 -\bg{\Omega}\odot (\bm{H}^{(\ell)}\bm{W}_2) + (1-\alpha) \bg{\Omega} \odot (\hat{\bm{A}} \bm{H}^{(\ell)}\bm{W}_2)
+\alpha\bg{\Omega} \odot (\bm{X} \bm{W}_3),
\label{Eq:weighted_jacobi}
\end{aligned}
\end{equation}
where $\odot$ denotes element-wise matrix multiplication, $\bm{W}_1$, $\bm{W}_2$, $\bm{W}_3$ are learnable weight matrices, and $\bg{\Omega}$ is a learnable weight modulation matrix. Notice that unlike the weighted Jacobi iteration, the proposed weighted Jacobi layer-wise propagation rule updates node features across layers, employs trainable weight matrices to learn an optimized graph representation, incorporates a learnable weight modulation matrix that functions similarly to the weighted parameter in the weighted Jacobi method, and applies a nonlinear activation function to capture the nonlinearity of the graph structure.

\medskip\noindent\textbf{Adjacency Modulation.}\quad The graph structure has a limitation in that it cannot capture relationships between distant nodes. To tackle this issue, we use adjacency modulation~\cite{zou2021modulated}, defined as $\check{\bm{A}}= \hat{\bm{A}} + \bm{Q}$, where $\bm{Q}$ is an $N\times N$ learnable modulation matrix.

\subsection{MLP-Graph Weighted Jacobi Mixer Model}
\begin{figure}[!htb]
\begin{center}
\includegraphics[scale=.78]{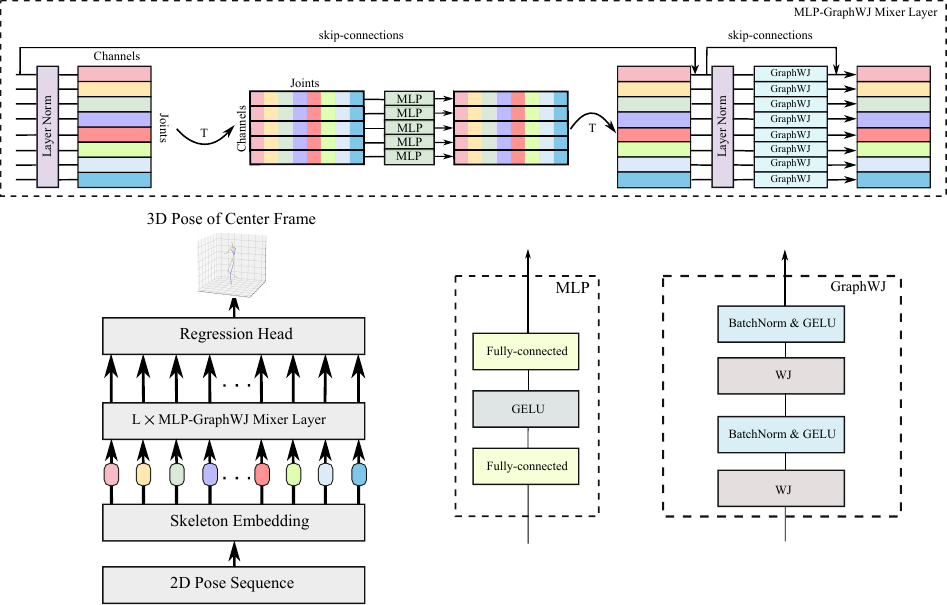}
\end{center}
\caption{Schematic diagram of the proposed network architecture for 3D human pose estimation. The architecture is comprised of three main components: skeleton embedding, MLP-GraphWJ mixer layer, and a regression head. The MLP-GraphWJ mixer layer consists of a joints mixing MLP layer and a GraphWJ mixing layer. The architecture also includes additional components such as skip connections, dropout, layer normalization, and batch normalization.}
\label{Fig:model}
\end{figure}

\noindent\textbf{Model Architecture.}\quad Inspired by the MLP-Mixer~\cite{tolstikhin2021mlp} and its recent variants for 3D human pose estimation and human motion forecasting tasks~\cite{Bouazizi:22,li2022graphmlp}, the architecture of the proposed MLP-GraphWJ mixer consists of three main stages: \textbf{\textsf{\footnotesize 1)}} skeleton embedding, \textbf{\textsf{\footnotesize 2)}} MLP-GraphWJ mixer layer, and \textbf{\textsf{\footnotesize 3)}} regression head. The overall architecture of the proposed model is illustrated in Figure~\ref{Fig:model}, which shows that the joint-mixing layer aggregates information across different positions within each channel using MLPs, while the GraphWJ mixing layer is responsible for aggregating information across different channels of the input using the weighted Jacobi (WJ) feature propagation rule. The output of the final GraphWJ mixing layer is then passed on to the regression head network.

\smallskip\noindent\textbf{\textsf{\small 1) Skeleton Embedding:}} To incorporate temporal information into our model, we take a 2D pose sequence as input. Given a 2D pose sequence $\bm{S}\in \mathbb{R}^{N\times 2\times T}$ represented as a tensor, where $T$ denotes the number of frames and $N$ is the number of joints, we first reshape it into a matrix $\tilde{\bm{S}}\in \mathbb{R}^{N \times 2T}$ by concatenating the 2D coordinates of all frames. Then, we pass it through a fully-connected layer, resulting in an $N\times F$ embedding matrix $\bm{X} = \tilde{\bm{S}}\bm{W}_{4}$, where $\bm{W}_4 \in \mathbb{R}^{2T \times F}$ is a learnable weight matrix and $F$ is the embedding dimension.

\smallskip\noindent\textbf{\textsf{\small 2) MLP-GraphWJ Mixer Layer:}} MLP-based models are not well-suited for handling graph-structured data, as they simply connect all nodes without considering the graph structure. To address this issue, we propose the MLP-GraphWJ mixer layer, which takes the advantages of both MLPs and graph neural networks in a single layer. Compared to the MLP-Mixer, our proposed MLP-GraphWJ mixer layer leverages graph neural networks to extract features of different channels, thereby helping to preserve domain-specific knowledge pertaining to human body configurations. Specifically, our MLP-GraphWJ mixer layer consists of two sub-layers: a joint-mixing MLP and a GraphWJ mixing layer. The joint-mixing MLP block allows communication between different joints, while the GraphWJ mixing layer allows communication between different channels. The joint-mixing MLP acts on the columns of the input feature matrix $\bm{H}^{(\ell)} \in \mathbb{R}^{N \times F}$ (i.e., applied to its transpose). On the other hand, the GraphWJ mixing layer acts on the rows of its input feature. The joint-mixing MLP block contains two fully-connected layers. We also add a skip connection between the input and output. Hence, the output of the joint-mixing MLP is an $N\times F$ matrix given by
\begin{equation}
\bm{U}^{(\ell+1)}=\bm{H}^{(\ell)}+\left(\bm{W}_6 \sigma\left(\bm{W}_5\left(\mathsf{LN}\left(\bm{H}^{(\ell)}\right)^{{\T}}\right)\right)\right)^{{\T}},
\label{Eq:spatial_mlp_3}
\end{equation}
where $\mathsf{LN}(\cdot)$ is layer normalization~\cite{ba2016layer}, $\bm{W}_5 \in \mathbb{R}^{N \times F}$ and $\bm{W}_6 \in \mathbb{R}^{F \times N}$ are learnable weight matrices. The input of the first layer is the $N\times F$ embedding matrix $\bm{H}^{(0)}=\bm{X}$.

On the other hand, our GraphWJ mixing layer consists of two weighted Jacobi ($\mathsf{WJ}$) layers. The output $\mathbf{U}^{(\ell+1)}$ of the joint-mixing MLP layer is fed into the GraphWJ mixing layer, which acts on the rows of its input matrix. Hence, the outputs of the first and second $\mathsf{WJ}$ layers are given by
\begin{equation}
\bm{P}^{(\ell+1)}=\sigma \Bigl(\mathsf{BN} \Bigl(\mathsf{WJ}\Bigl( \mathbf{U}^{(\ell+1)} \Bigl)\Bigl)\Bigl) \in \mathbb{R}^{N\times R}
\label{Eq:wj_pl_1}
\end{equation}
and
\begin{equation}
\bm{Q}^{(\ell+1)}=\sigma\Bigl(\mathsf{BN} \Bigl(\mathsf{WJ}\Bigl(\bm{P}^{(\ell+1)}\Bigl) \Bigl)\Bigl)  \in \mathbb{R}^{N \times F},
\label{Eq:wj_ql_1}
\end{equation}
where $\mathsf{BN}(\cdot)$ is a batch normalization layer, and $R$ and $F$ are embedding dimensions. Batch normalization is similar to layer normalization, but instead of normalizing across the features of each input, it normalizes across a batch of inputs.

Finally, the output $\bm{Z}$ of the last MLP-GraphWJ mixing layer is obtained by adding a skip connection as follows:
\begin{equation}
\bm{Z}= \bm{U}^{(L)} + \bm{Q}^{(L)} \in \mathbb{R}^{N\times F}
\label{Eq:wj_hl}
\end{equation}

\smallskip\noindent\textbf{\textsf{\small 3) Regression Head:}} The output $\bm{Z}$ of the last MLP-GraphWJ mixing layer is passed on to the regression head network comprised of a layer normalization, followed by a linear fully connected layer, yielding a prediction $\hat{\bm{Y}}=(\hat{\bm{y}}_{1},\dots,\hat{\bm{y}}_{N})^{\T}\in \mathbb{R}^{N \times 3}$ of estimated 3D joint positions. This prediction is regressed to the ground-truth of the 3D pose for the center frame during model training.

\medskip\noindent\textbf{Model Training.}\quad In order to train the MLP-GraphWJ mixer model for 3D human pose estimation, the weight matrices for various layers are optimized by minimizing the following loss function
\begin{equation}
\mathcal{L} =\frac{1}{N}\left[(1-\lambda)\sum_{i=1}^{N}\Vert\bm{y}_{i}-\hat{\bm{y}}_{i}\Vert_{2}^{2}+
\lambda\sum_{i=1}^{N}\Vert\bm{y}_{i}-\hat{\bm{y}}_{i}\Vert_{1}\right],
\end{equation}
which is a weighted combination of the mean squared and mean absolute errors between the estimated 3D joint positions $\hat{\bm{y}}_{i}$ and the ground-truth positions $\bm{y}_{i}$ over $N$ training body joints.

\section{Experiments}
In this section, we evaluate the performance of our model against competitive baselines for 3D human pose estimation. More detailed descriptions of the datasets, additional experimental results and ablation studies are provided in the supplementary material. Code is available at: \textcolor{blue}{https://github.com/nies14/Spatio-Temporal-MLP-Graph}
\subsection{Experimental Setup}
\noindent\textbf{Datasets.}\quad We assess the performance of our model on two widely used benchmark datasets for 3D human pose estimation: Human3.6M~\cite{ionescu2013human3} and MPI-INF-3DHP~\cite{Dushyant:2017}.

\medskip\noindent\textbf{Evaluation Protocols and Metrics.}\quad For Human3.6M, we adopt two commonly used metrics, mean per joint position error (MPJPE) and Procrustes-aligned mean per joint position error (PA-MPJPE), which are measured in millimeters. A lower value of these metrics indicates better performance. For MPI-INF-3DHP, we evaluate our model using two standard metrics: Percentage of Correct Keypoints (PCK) within 150mm and Area Under the Curve (AUC), consistent with previous studies~\cite{quan2021higher,zou2021modulated,yang20183d,pavlakos2018ordinal,Habibie:19,li2019generating}. Improved model performance is indicated by higher values of PCK and AUC.

\medskip\noindent\textbf{Baselines.}\quad We evaluate the performance of our MLP-GraphWJ mixer model against various state-of-art methods, including semantic GCN~\cite{zhao2019semantic}, spatio-temporal GCN (ST-GCN)~\cite{YujunCai:19}, Weight Unsharing~\cite{liu2020comprehensive}, temporal convolutions and semi-supervised training~\cite{pavllo20193d}, skeletal GNN~\cite{zeng2021learning}, graph mixture density network (GraphMDN)~\cite{oikarinen2021graphmdn}, split-and-recombine network (SRNet)~\cite{Zeng2020SRNet}, graph attention spatio-temporal network (GAST-Net)~\cite{liu2021graph}, PoseFormer~\cite{PoseFormer:2021}, modulated GCN (MGCN)~\cite{zou2021modulated}, group graph convolutional networks (GroupGCN)~\cite{groupgraph2022}, mesh transformer (METRO)~\cite{lin2021end}, and pose augmentation (PoseAug)~\cite{gong2021poseaug}.

\medskip\noindent\textbf{Implementation Details.}\quad We train the model using AMSGrad optimizer for 50 epochs, and the initial learning rate is set to 0.001 with a decay factor of 0.95 applied after each epoch and 0.5 after every 5 epochs. For 2D pose detections~\cite{chen2018cascaded}, we set the batch size to 256, the number of layers $L=3$, the skeleton embedding layer hidden dimension and the MLP hidden dimension $F=384$, and the GraphWJ mixing layer hidden dimension $R=768$. We set the weighting factor $\lambda=0.01$, $\alpha=0.1$, and the total number of input frames $T=243$ for both 2D detected poses and ground truth poses.

\subsection{Results and Analysis}
\noindent\textbf{Quantitative Results.}\quad In Table~\ref{Tab:Result1}, we report the performance comparison results of our MLP-GraphWJ mixer model and various state-of-art methods for 3D human pose estimation. As can be seen, our model demonstrates superior performance with detected 2D pose as an input across most actions and overall, as evidenced by both Protocol \#1 and Protocol \#2. These findings demonstrate the model's competitiveness, which is largely attributed to the fact that MLP-GraphWJ mixer can better exploit joint connections through the proposed graph propagation rule and also learns not only different modulation vectors for different body joints, but also additional connections between the joints. Under Protocol \#1, Table~\ref{Tab:Result1} shows that using a single frame MLP-GraphWJ mixer performs better than MGCN~\cite{zou2021modulated} on 14 out of 15 actions by a relative improvement of 10.73\% on average. Of significance is the fact that unlike our method, MGCN~\cite{zou2021modulated} employs a non-local layer. Despite this difference, our model demonstrates superior performance compared to MGCN~\cite{zou2021modulated}, highlighting the efficacy of our approach. Our model also performs better than Skeletal GNN~\cite{zeng2021learning}, a temporal graph neural network method for hard 3D pose estimation, yielding an error reduction of approximately 3.92\% on average. Under Protocol \#2, our approach outperforms spatio-temporal GCN with a relative improvement (average) of 6.67\% in terms of PA-MPJPE.

\begin{table}[!h]
\caption{Performance comparison of our model and baseline methods on Human3.6M under Protocol \#1 and Protocol \#2 using the detected 2D pose as input. The average errors are reported in the last column. Boldface numbers indicate the best performance, and the underlined numbers indicate the second-best performance. ($\dagger$) - uses temporal information.}
\footnotesize
\setlength{\tabcolsep}{.4pt}
\smallskip
\centering
\begin{tabular}{l*{17}{c}}
\toprule[1pt]
& \multicolumn{15}{c}{Action}\\
\cmidrule(lr){2-16}
\textbf{Protocol \#1} & Dire. & Disc. &  Eat & Greet & Phone & Photo &  Pose & Purch. & Sit & SitD. & Smoke & Wait & WalkD. & Walk & WalkT. & Avg.\\
\midrule[.8pt]
Zhao \etal~\cite{zhao2019semantic} & 47.3& 60.7& 51.4 &60.5& 61.1& 49.9 & 47.3& 68.1 &86.2& \textbf{55.0}& 67.8& 61.0& \textbf{42.1}& 60.6& 45.3& 57.6\\
Quan \etal~\cite{quan2021higher} & 47.0 & 53.7 & 50.9 & 52.4&57.8 &71.3&50.2 &49.1 &63.5 &76.3  &54.1&51.6 &56.5 &41.7 &45.3 & 54.8 \\
Liu \etal~\cite{liu2020comprehensive} & 46.3 & 52.2 & 47.3 & 50.7 & 55.5 & 67.1 & 49.2 & 46.0 & 60.4 & 71.1 & 51.5 & 50.1 & 54.5 & 40.3 & 43.7 & 52.4 \\
Lin \etal~\cite{lin2021end} &  - & - & - & - & - & - & - & - & - & - & - & - & - & - & - & 54.0 \\
Zhao \etal~\cite{zhao2022graformer} & 45.2 & 50.8 & 48.0 & 50.0 & 54.9 & 65.0 & 48.2 & 47.1 & 60.2 & 70.0 & 51.6 & 48.7 & 54.1 & 39.7 & 43.1 & 51.8 \\
Lee \etal~\cite{multihop2022} & 46.8 & 51.4 & 46.7 & 51.4 & 52.5 & 59.7 & 50.4 & 48.1 & 58.0 & 67.7 & 51.5 & 48.6 & 54.9 & 40.5 & 42.2 & 51.7 \\
Zhang~\cite{groupgraph2022} & 45.0 & 50.9 & 49.0 & 49.8 & 52.2 & 60.9 & 49.1 & 46.8 & 61.2 & 70.2 & 51.8 & 48.6 & 54.6 & 39.6 & 41.2 & 51.6 \\
Gong \etal~\cite{gong2021poseaug} & - & - & - & - & - & - & - & - & - & - & - & - & - & - & - & 50.2 \\
Zou \etal~\cite{zou2021modulated} & 45.4 & 49.2 & 45.7 & 49.4 & 50.4 &  58.2 & 47.9 & 46.0 & 57.5 & 63.0 & 49.7 & 46.6 & 52.2 & 38.9 & 40.8 & 49.4\\
Cai \etal~\cite{YujunCai:19} ($\dagger$) & 44.6 & 	47.4 & 45.6 & 48.8 & 50.8 & 59.0 & 47.2 & 43.9 & 57.9 & 61.9 & 49.7 & 46.6 & 51.3 & 37.1 & 39.4 & 48.8\\
Li \etal~\cite{li2022graphmlp} & 43.7 & 49.3 & 45.5 & 47.8 & 50.5 & 56.0 & 46.3 & 44.1 & 55.9 & 59.0 & 48.4 & 45.7 & 51.2 & 37.1 & 39.1 & 48.0 \\
Pavllo \etal~\cite{pavllo20193d} ($\dagger$) & 45.2 &  46.7 & 43.3 & 45.6 & 48.1 & 55.1 & 44.6 & 44.3 & 57.3 & 65.8 & 47.1 & 44.0 & 49.0 & 32.8 & 33.9 & 46.8 \\
Oikarinen \etal~\cite{oikarinen2021graphmdn} & \underline{40.0} &  \textbf{43.2} & 41.0 & 43.4 & 50.0 & 53.6 &  \textbf{40.1} & 41.4 &  \textbf{52.6} & 67.3 & 48.1 & 44.2 & 49.0 & 39.5 & 40.2 & 46.2 \\
Zeng \etal~\cite{zeng2021learning} ($\dagger$)  & 43.1 & 50.4 & 43.9 & 45.3 & 46.1 & 57.0 & 46.3 & 47.6 & 56.3 & 61.5 & 47.7 & 47.4 & 53.5 & 35.4 & 37.3 & 47.9 \\
Zeng \etal~\cite{Zeng2020SRNet} ($\dagger$) & 46.6 & 47.1 & 43.9 &  \textbf{41.6} & \underline{45.8} & \underline{49.6} & 46.5 & \underline{40.0} & 53.4 & 61.1 & 46.1 &  \textbf{42.6} &  \underline{43.1} & \underline{31.5} & 32.6 & 44.8 \\
Liu \etal~\cite{liu2021graph} ($\dagger$) & 43.3 & 46.1 &  \underline{40.9} & 44.6 & 46.6 & 54.0 & 44.1 & 42.9 & 55.3 & \underline{57.9} & 45.8 & 43.4 & 47.3 &  \textbf{30.4} & \textbf{30.3} & 44.9 \\
Zheng \etal~\cite{PoseFormer:2021} ($\dagger$) & 41.5 & 44.8 &  \textbf{39.8} & \underline{42.5} & 46.5 & 51.6 & \underline{42.1} & 42.0 & \underline{53.3} & 60.7 & \underline{45.5} & 43.3 & 46.1 & 31.8 & \underline{32.2} & \underline{44.3}\\
\midrule[.8pt]
Ours ($\dagger$) & \textbf{38.9} & \underline{44.5} & 41.4 & 43.7 & \textbf{45.0} &  \textbf{48.7} & 42.8 &  \textbf{39.5} & 54.9 & 67.1 &  \textbf{42.5} & \underline{43.1} & 44.0 & 33.2 & 33.0 &  \textbf{44.1}\\
\midrule[1pt]
\textbf{Protocol \#2} & Dire. & Disc. &  Eat & Greet & Phone & Photo &  Pose & Purch. & Sit & SitD. & Smoke & Wait & WalkD. & Walk & WalkT. & Avg.\\
\midrule[.8pt]
Lee \etal~\cite{Lee2018LSTM} ($\dagger$)  & 34.9 & 35.2 & 43.2 & 42.6 & 46.2 & 55.0 & 37.6 & 38.8 & 50.9 & 67.3 & 48.9 & 35.2 & 31.0 & 50.7 & 34.6 & 43.4 \\
Quan \etal~\cite{quan2021higher} & 36.9 & 42.1&40.3 &42.1 &43.7 &52.7&37.9 &37.7 &51.5 &60.3  &43.9&39.4 & 45.4 & 31.9 & 37.8 & 42.9 \\
Liu  \etal~\cite{liu2020comprehensive} & 35.9 & 40.0 & 38.0 & 41.5 & 42.5 & 51.4 & 37.8 & 36.0 & 48.6 & 56.6 & 41.8 & 38.3 & 42.7 & 31.7 & 36.2 & 41.2\\
Lee \etal~\cite{multihop2022} & 35.7 & 39.6 & 37.3 & 41.4 & 40.0 & 44.9 & 37.6 & 36.1 & 46.5 & 54.1 & 40.9 & 36.4 & 42.8 & 31.7 & 34.7 & 40.3 \\
Zhang~\cite{groupgraph2022}  & 35.3 & 39.3 & 38.4 & 40.8 & 41.4 & 45.7 & 36.9 & 35.1 & 48.9 & 55.2 & 41.2 & 36.3 & 42.6 & 30.9 & 33.7 & 40.1 \\
Zou \etal~\cite{zou2021modulated} & 35.7 & 38.6 & 36.3 & 40.5 &  \underline{39.2} & 44.5 & 37.0 & 35.4 & 46.4 &  51.2 & 40.5 &35.6 & 41.7 & 30.7 & 33.9 & 39.1\\
Gong \etal~\cite{gong2021poseaug} & - & - & - & - & - & - & - & - & - & - & - & - & - & - & - & 39.1 \\
Cai \etal~\cite{YujunCai:19} ($\dagger$) & 35.7 & 37.8 & 36.9 & 40.7 & 39.6 & 45.2 & 37.4 & 34.5 & 46.9 & \textbf{50.1} & 40.5 & 36.1 & 41.0 & 29.6 & 33.2 & 39.0\\
Lin \etal~\cite{lin2021end}  & - & - & - & - & - & - & - & - & - & - & - & - & - & - & - & 36.7 \\
Pavllo \etal~\cite{pavllo20193d} ($\dagger$) &  \underline{34.1} & \textbf{36.1} & \underline{34.4} & \textbf{37.2} & \textbf{36.4} &  \underline{42.2} &  \underline{34.4} &  \underline{33.6} & \textbf{45.0} &  \underline{52.5} &  \underline{37.4} &  \underline{33.8} &  \underline{37.8} & \textbf{25.6} &  \underline{27.3} &  \underline{36.5} \\
\midrule[.8pt]	
Ours ($\dagger$) & \textbf{33.0} & \underline{36.8} & \textbf{34.3} & \underline{37.5} & \textbf{36.4} & \textbf{40.4} & \textbf{34.1} & \textbf{31.9} &  \underline{45.4} & 57.0 & \textbf{35.6} & \textbf{34.8} & \textbf{36.2} &  \underline{26.5} & \textbf{26.9} & \textbf{36.4}\\
	
\bottomrule[1pt]
\end{tabular}
\label{Tab:Result1}
\end{table}

\noindent In Table~\ref{Tab:MPI}, we report the quantitative comparison results of MLP-GraphWJ mixer using a single frame in comparison with several baselines on the MPI-INF-3DHP dataset. As can be seen, our method achieves significant improvements over the comparative methods. Our model outperforms the best performing baseline with relative improvements of 0.81\% and 1.30\% in terms of the PCK and AUC metrics, respectively. Although we train the model using only the Human3.6M dataset, our method outperforms others on MPI-INF-3DHP, indicating that our approach has strong generalization capabilities to unseen human poses.

\begin{table}[!h]
\caption{Performance comparison of our model without pose refinement and baseline methods on the MPI-INF-3DHP dataset using PCK and AUC as evaluation metrics.}
\small
\setlength{\tabcolsep}{2.5pt}
\smallskip
\centering
\begin{tabular}{lcc}
\toprule[1pt]
Method & PCK($\uparrow$) & AUC($\uparrow$)\\
\midrule[.8pt]
Chen \etal~\cite{li2019generating} & 67.9 & - \\
Yang \etal~\cite{yang20183d} & 69.0 & 32.0 \\
Pavlakos \etal~\cite{pavlakos2018ordinal}  & 71.9 & 35.3 \\
Habibie \etal~\cite{Habibie:19}  & 70.4 & 36.0 \\
Quan \etal~\cite{quan2021higher} & 72.8 &36.5 \\
Zeng \etal~\cite{Zeng2020SRNet} & 77.6 & 43.8\\
Zhang \etal~\cite{groupgraph2022} & 81.1 & 49.9\\
Zeng \etal~\cite{zeng2021learning} & 82.1 & 46.2\\
Zou \etal~\cite{zou2021modulated} & 86.1 & 53.7\\
\midrule[.8pt]
Ours & \textbf{86.8} &\textbf{54.4} \\
\bottomrule[1pt]
\end{tabular}
\label{Tab:MPI}
\end{table}

\medskip\noindent\textbf{Qualitative Results.}\quad Figure~\ref{Fig:visual} shows some visualization results of the proposed MLP-GraphWJ mixer model on Human3.6M. As can be seen, the 3D predictions on various actions made by our model are superior to those of MGCN~\cite{zou2021modulated} and more closely match the ground truth. This indicates the effectiveness of our approach. Notice that MGCN~\cite{zou2021modulated} struggles to accurately predict hand poses when there are overlapping joints or occlusions, whereas our model is able to predict them with a high degree of accuracy.
\begin{figure}[!htb]
\centering
\includegraphics[scale=1.1]{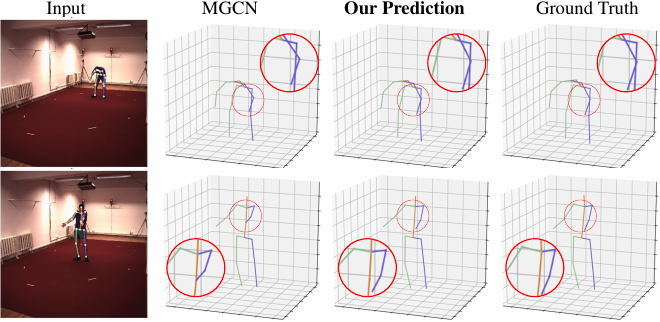}
\caption{Qualitative comparison between our model and MGCN on Human3.6M actions.}
\label{Fig:visual}
\end{figure}

\subsection{Ablation Study}
We conduct ablation experiments on the Human3.6M dataset under Protocol \#1 using MPJPE as evaluation metric. Specifically, we investigate the effectiveness of each component in our network architecture. The results are presented in Table~\ref{Tab:eachComponents}, where the first row corresponds to the performance of the MLP-Mixer baseline model~\cite{tolstikhin2021mlp} that does not include any GCN components. The remaining rows in the table display the results of replacing various components of the baseline model. We fix the number of parameters to be about 0.95M by merely changing the number of hidden dimensions of each model. Our proposed MLP-GraphWJ mixer outperforms the baseline model by a margin of 1.9mm, demonstrating that the combined use of these components leads to more accurate 3D pose estimation.

\begin{table}[!htb]
\caption{Effectiveness of each component used in our method without pose refinement on Human3.6M under Protocol\#1 using detected 2D poses as input.}
\smallskip
\centering
\begin{tabular}{cccccc}
\toprule[1pt]
\parbox[t]{1.0cm}{\centering  \small{Joint-Mixing} \\  \small{MLP}} &  \parbox[t]{1.1cm}{\centering Channel-Mixing \\ MLP} & \parbox[t]{1.3cm}{\centering GraphWJ Mixing \\ Layer} & \parbox[t]{1.0cm}{\centering Standard \\ GCN} & \parbox[t]{1.3cm}{\centering Weighted \\ Jacobi} &  MPJPE ($\downarrow$) \\
\midrule[.8pt]
\cmark & \cmark & \xmark & \xmark & \xmark & 53.1 \\
\cmark & \xmark & \cmark & \cmark & \xmark & 51.5\\
\cmark & \xmark & \cmark & \xmark & \cmark & \textbf{51.2} \\
\bottomrule[1pt]
   \end{tabular}
\label{Tab:eachComponents}
\end{table}

\medskip\noindent\textbf{Runtime Analysis.}\quad We report the model performance, the total number of parameters, and estimated floating-point operations (FLOPs) per frame with various input sequence lengths ($T$) in Table~\ref{Tab:complexity}. We can see that increasing the sequence length of our model results in improved accuracy, while keeping the total number of learned parameters low.

\begin{table}[!htb]
\caption{Comparison of our model and baselines in terms of total number of parameters, FLOPs, and MPJPE. The evaluation is performed without pose refinement on Human3.6M under Protocol\#1 using detected 2D poses as input.}
\setlength{\tabcolsep}{4pt}
\smallskip
\centering
\begin{tabular}{lrrrc}
\toprule[1pt]
Method & Frames &  Params. & FLOPs & MPJPE($\downarrow$) \\
           & $(T)$ & $(\text{M})$ & $(\text{M})$ & \\
\midrule[.8pt]
VideoPose~\cite{pavllo20193d} & 27 & 8.56 & 17.09 & 48.8 \\
PoseFormer~\cite{PoseFormer:2021} & 9 & 9.58 & 150.0 & 49.9 \\
Ray3D~\cite{Zhan_2022_CVPR} & 9 & 27.50 & - & 49.7 \\
\midrule[.8pt]
Ours & 1 & 5.42 & 29.01 & 50.8 \\
Ours & 9 & 5.43 & 29.21 & \textbf{48.7} \\
\bottomrule[1pt]
\end{tabular}
\label{Tab:complexity}
\end{table}


\section{Conclusion}
In this paper, we proposed a novel network architecture, named MLP-GraphWJ mixer, which is comprised of an MLP-mixer layer and a GraphWJ mixer layer. The MLP-mixer layer aggregates information across different positions within each channel, while the graph weighted Jacobi network layer aggregates information across different channels. Our contribution involves the incorporation of a feature propagation rule based on weighted Jacobi iteration, which is derived through the application of graph filtering with implicit fairing. The proposed model outperforms recent state-of-the-art techniques on two standard benchmark datasets for 3D human pose estimation, as demonstrated by our experimental results. Our approach achieves this improved performance while employing a model with a smaller parameter count. For future work, we aim to take high-order connectivity between joints into account by aggregating information from multi-hop neighbors.

\medskip\noindent\textbf{Acknowledgments.}\quad This work was supported in part by the Discovery Grants program of Natural Sciences and Engineering Research Council of Canada.

\bibliographystyle{bmvc2k_natbib}
\bibliography{references}

\clearpage
\setcounter{page}{1}
\section*{----- Supplementary Material -----}
This supplementary material includes more detailed descriptions of the datasets, and additional experimental results.

\subsection*{Datasets and Implementation Details}
\noindent\textbf{Human3.6M} is a large-scale dataset containing more than 3.6 million human poses, and includes 15 different human activities performed by 11 actors~\cite{ionescu2013human3}. During training, we use 5 subjects (S1, S5, S6, S7, S8), and during testing, we use 2 subjects (S9, S11) from the dataset.

\medskip\noindent\textbf{MPI-INF-3DHP} contains 1.3 million frames and features 8 actors performing 8 actions, providing a wider range of poses~\cite{Dushyant:2017}. It includes a test set with 6 subjects in both indoor and complex outdoor scenes, enabling the evaluation of the model's generalization ability to unseen environments.

\medskip\noindent\textbf{More Implementaion Details.}\quad All experiments are conducted on a single NVIDIA GeForce RTX 3070 GPU with 8G memory, and our model is implemented in PyTorch. For the 2D ground truth, we set the batch size to 256, $L=3$, $F=128$, and $R=256$. To prevent overfitting, we also add dropout with a factor of 0.2 after each graph weighted Jacobi layer.

\section{Additional Experimental Results}

\noindent\textbf{Quantitative Results.}\quad Table~\ref{Tab:Result2} reports the results of our MLP-GraphWJ mixer model and various competing baselines when using 2D ground truth keypoints as input. The findings indicate that our model outperforms GraphMDN~\cite{oikarinen2021graphmdn} on 12 out of 15 actions with an average error reduction of approximately 2.42\% under Protocol \#1. Moreover, our model shows better performance compared to MGCN~\cite{zou2021modulated}, High-Order GCN~\cite{zou2020high}, SemGCN~\cite{zhao2019semantic}, and Weight Unsharing~\cite{liu2020comprehensive} on average, while having a lower number of learnable parameters and inference time. These results highlight the effectiveness of our proposed method.

\begin{table}[!h]
\caption{Performance comparison of our model and baseline methods on Human3.6M under protocol \#1 using the ground truth 2D pose as input. Boldface numbers indicate the best performance, whereas the underlined numbers indicate the second-best performance. ($\dagger$) - uses temporal information.}
\footnotesize
\setlength{\tabcolsep}{.4pt}
\smallskip
\centering
\begin{tabular}{l*{17}{c}}
\toprule[1pt]
& \multicolumn{15}{c}{Action}\\
\cmidrule(lr){2-16}
\textbf{Protocol \#1} & Dire. & Disc. &  Eat & Greet & Phone & Photo &  Pose & Purch. & Sit & SitD. & Smoke & Wait & WalkD. & Walk & WalkT. & Avg.\\
\midrule[.8pt]
Martinez \textit{et al.}~\cite{martinez2017simple} & 37.7 & 44.4 & 40.3 & 42.1 & 48.2 & 54.9 & 44.4 & 42.1 & 54.6 & 58.0 & 45.1 & 46.4 & 47.6 & 36.4 & 40.4 & 45.5\\
Pavlakos \textit{et al.}~\cite{pavlakos2018ordinal} & 47.5 & 50.5 & 48.3 & 49.3 & 50.7 & 55.2 & 46.1 &  48.0 & 61.1 & 78.1 & 51.1 & 48.3 & 52.9 & 41.5 & 46.4 & 51.9\\
Hossain \textit{et al.}~\cite{hossain2018exploiting} ($\dagger$) & 35.7 & 39.3 & 44.6 & 43.0 & 47.2 & 54.0 & 38.3 & 37.5 & 51.6 & 61.3 & 46.5 & 41.4 & 47.3 & 34.2 & 39. & 44.1 \\
Cai \textit{et al.}~\cite{YujunCai:19} ($\dagger$) & 32.9 & 38.7 & 32.9 & 37.0 & 37.3 & 44.8 & 38.7 & 36.1 & 41.0 & 45.6 & 36.8 & 37.7 & 37.7 & 29.5 & 31.6 & 37.2\\
Liu \textit{et al.} \cite{liu2020comprehensive} & 36.8 & 40.3 & 33.0 & 36.3 & 37.5 & 45.0 & 39.7 & 34.9 & 40.3 & 47.7 & 37.4 & 38.5 & 38.6 & 29.6 & 32.0 & 37.8 \\
Pavllo \textit{et al.}~\cite{pavllo20193d} ($\dagger$)  & 35.2 & 40.2 & 32.7 & 35.7 & 38.2 & 45.5 & 40.6 & 36.1 & 48.8 & 47.3 & 37.8 & 39.7 & 38.7 & 27.8 & 29.5 & 37.8 \\
Zou \textit{et al.} \cite{zou2021modulated} & - & - & - & - & - & - & - & - & - & - & - & - & - & - & - & 37.4 \\
Oikarinen \textit{et al.}~\cite{oikarinen2021graphmdn} & 33.9 & 39.9 & 33.0 & 35.4 & 36.8 & 44.4 & 38.9 & 33.0 & 41.0 & 50.0 & 36.4 & 38.3 & 37.8 & 28.2 & 31.5 & 37.2 \\
Lee \textit{et al.}~\cite{multihop2022} & 34.6 & 39.6 & \underline{31.3} & 34.7 & \underline{33.9} & 40.3 & 39.5 & 32.2 & \textbf{35.4} & 43.5 & \underline{34.0} & \underline{35.0} & 36.9 & 29.7 & 31.4 & 35.6 \\
Zhang \textit{et al.}~\cite{groupgraph2022} &  - & - & - & - & - & - & - & - & - & - & - & - & - & - & - & 35.3 \\
Zhao \textit{et al.}~\cite{zhao2022graformer} & 32.0 & 38.0 & \textbf{30.0} & 34.4 & 34.7 & 43.3 & \textbf{35.2} & 31.4 & \underline{38.0} & 46.2 & 34.2 & 35.7 & 36.1 & \underline{27.4} & 30.6 & 35.2 \\
Zhan \textit{et al.}~\cite{Zhan_2022_CVPR} ($\dagger$) & \textbf{31.2} & \underline{35.7} & 31.4 & \underline{33.6} & 35.0 & \underline{37.5} & 37.2 & \underline{30.9} & 42.5 & \textbf{41.3} & 34.6 & 36.5 & \underline{32.0} & 27.7 & \underline{28.9} & \underline{34.4} \\
\midrule[.8pt]
Ours ($\dagger$) & \underline{31.6} & \textbf{35.6} & 31.5 & \textbf{31.0} & \textbf{32.1} & \textbf{35.1} & \underline{36.3} & \textbf{30.1} & 38.8 & \underline{41.4} & \textbf{32.6} & \textbf{34.6} & \textbf{31.4} & \textbf{25.5} & \textbf{25.8} & \textbf{32.9} \\
\bottomrule[1pt]
\end{tabular}
\label{Tab:Result2}
\end{table}

\medskip\noindent\textbf{Qualitative Results.}\quad Figure~\ref{Fig:visual} shows some additional visualization results of the proposed MLP-GraphWJ mixer model on the Human3.6M dataset. Our model demonstrates a high degree of accuracy in predicting hand poses, even in scenarios where joints overlap or occlusions occur, while MGCN~\cite{zou2021modulated} struggles to perform the same task effectively.
\begin{figure}[!htb]
\centering
\setlength{\tabcolsep}{7pt}
\begin{tabular}{cccc}
\hspace*{.15in} Input & \hspace*{.55in} MGCN & \hspace*{.35in} \textbf{Our Prediction} & \hspace*{.25in} Ground Truth \\[-.65ex]
\end{tabular}
\includegraphics[scale=.78]{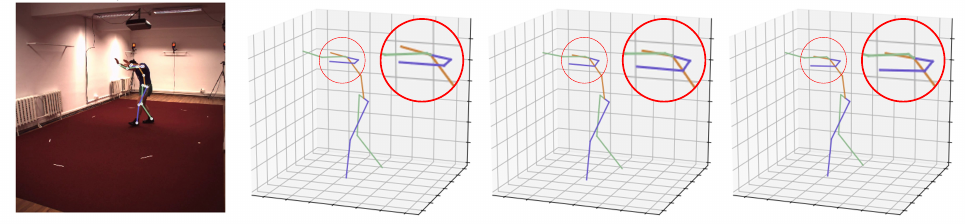} \\
\includegraphics[scale=.78]{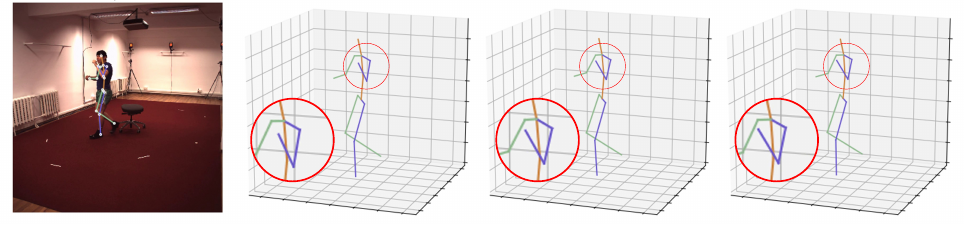}
\caption{Qualitative comparison between our model and MGCN on Human3.6M actions. The red circle indicates the locations where our model yields better results.}
\label{Fig:visual}
\end{figure}

\begin{figure}[!htb]
\begin{center}
\includegraphics[scale=.5]{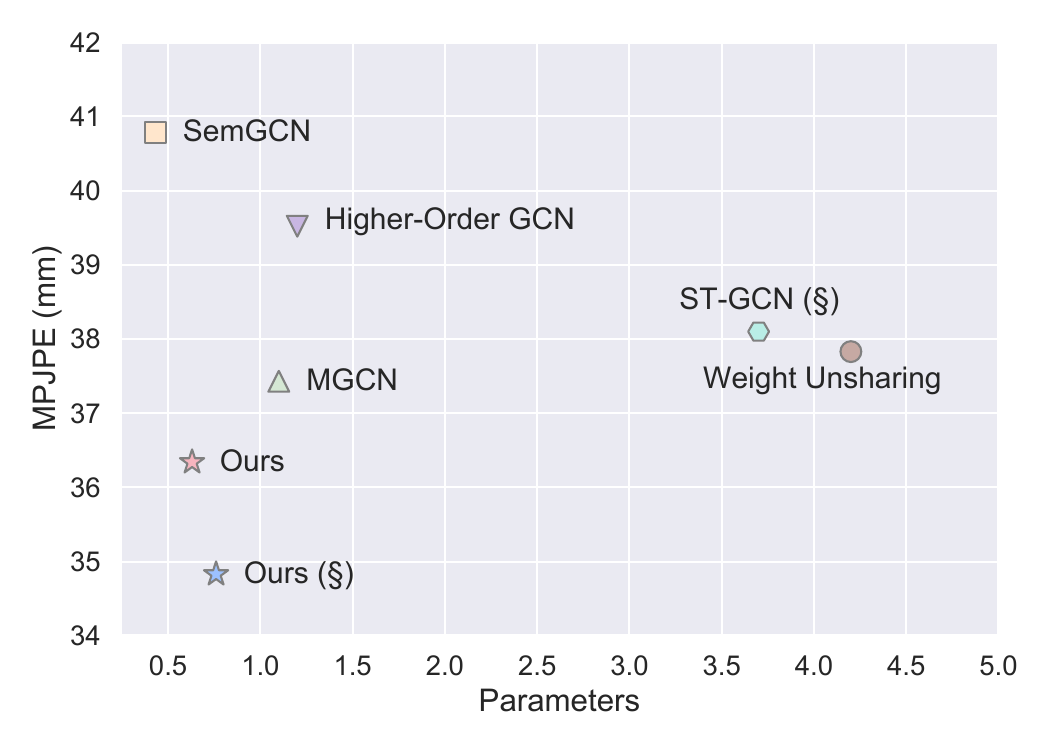}
\end{center}
\caption{Performance and model size comparison between our model and state-of-the-art methods for 3D human pose estimation, including MGCN~\cite{zou2021modulated}, SemGCN~\cite{zhao2019semantic}, High-Order GCN~\cite{zou2020high}, ST-GCN~\cite{YujunCai:19}, and Weight Unsharing~\cite{Kenkun:2020}. Lower Mean Per Joint Position Error (MPJPE) values indicate better performance. Evaluation conducted on a single frame of Human3.6M~\cite{ionescu2013human3} dataset with 2D joints as input. (\S) - uses a pose refinement network.}
\label{Fig:humangraph2}
\end{figure}

\medskip\noindent\textbf{Model Size Comparison.}\quad The proposed framework employs a weighted Jacobi (WJ) feature propagation rule obtained via graph filtering with implicit fairing. One of the key benefits of our model is that it presents a simple and competitive alternative to existing approaches that do not use self-attention mechanisms, while outperforming previous work and retaining a small model size, as illustrated in Figure~\ref{Fig:humangraph2}. Moreover, our approach effectively merges temporal information within the feature channels, while incurring minimal computational cost in terms of sequence length.

\medskip\noindent\textbf{Hyper-Parameter Sensitivity Analysis.}\quad We start by investigating the impact of the different hyper-parameters on model performance. Results are reported in Table~\ref{Tab:modelConfigurations}. It can be observed that the expanding ratio of 2 ($F=384$, $R=768$) performs better than the commonly used ratio of 4 in vision Transformers and MLPs. The value of the skeleton embedding hidden dimension $F$ affects the model ability to capture patterns. When increasing $F$ from 128 to 384 and $R$ from 256 to 768, the MPJPE decreases from 47.5mm to 45.3mm. However, the number of trainable parameters increases from 0.65M to 5.48M. The best results are obtained using $F=384$, and $R=768$. Using three MLP-GraphWJ mixer layers yields the best performance, while increasing or decreasing the number of layers negatively impacts performance.

\begin{table}[!htb]
\caption{Ablation study on various configurations of our approach without pose refinement on Human3.6M under protocol\#1 using detected 2D pose as input. $L$ is the number of MLP-GraphWJ mixer layers, $F$ is the hidden dimension of skeleton embedding and joints mixing MLP and $R$ is the hidden dimension of GraphWJ mixing layer. The number of input frames is set to $T=81$. Boldface numbers indicate the best performance.}
\small
\setlength{\tabcolsep}{6pt}
\smallskip
\centering
\begin{tabular}{crrcc}
\toprule[1pt]
$L$ & $F$ & $R$ & Params. $(\text{M})$ & MPJPE ($\downarrow$) \\
\midrule[.8pt]
3 & 128 & 256 & 0.65 & 47.5 \\
3 & 256 & 256 & 1.28 & 47.7 \\
3 & 256 & 512 & 2.47 & 47.9 \\
3 & 256 & 1024 & 4.86 & 47.3 \\
3 & 384 & 384 & 2.80 & 46.8 \\
3 & 384 & 768 & 5.48 & \textbf{45.3}  \\
3 & 384 & 1536 & 10.83 & 46.1 \\
1 & 384 & 768 & 1.87 & 48.3 \\
2 & 384 & 384 & 3.68 & 46.6 \\
4 & 384 & 768 & 7.29 & 46.6 \\
\bottomrule[1pt]
\end{tabular}
\label{Tab:modelConfigurations}
\end{table}

\medskip\noindent\textbf{Comparison with GCN-based Methods.}\quad In order to bypass the influence of 2D pose detectors and gain further insight into the importance of our network architecture and graph propagation rule, we train our model on the Human3.6M dataset using 2D ground truth poses by maintaining the expanding ratio of 2 ($F=128$, $R=256$) and we report the results in Table~\ref{Tab:groundTruth}. Our method demonstrates superior performance compared to recent state-of-art methods based on a single frame, despite utilizing fewer trainable parameters.
\begin{table}[!htb]
\caption{Performance comparison of our model and baseline methods without pose refinement using ground-truth keypoints. Boldface numbers indicate the best performance.}
\smallskip
\centering
\begin{tabular}{lrcccc}
\toprule[1pt]
Method & Filters &  Params & MPJPE & PA-MPJPE & Infer.\\
       &         & $(\text{M})$   & ($\downarrow$) & ($\downarrow$) & Time \\
\midrule[.8pt]
SemGCN~\cite{zhao2019semantic} & 128 & 0.43 & 40.78 & 31.46 & .012s \\
High-Order GCN~\cite{zou2020high}  & 96 & 1.20 & 39.52 &31.07 & .013s \\
Weight Unsharing~\cite{liu2020comprehensive} & 128 & 4.22 & 37.83 & 30.09 & .032s  \\
MGCN~\cite{zou2021modulated} & 256 & 1.10 & 37.43 & 29.73 & .008s \\
\midrule[.8pt]
Ours & - & 0.63 & \textbf{36.34} & \textbf{28.97} & .005s \\
\bottomrule[1pt]
\end{tabular}
\label{Tab:groundTruth}
\end{table}

\begin{figure}[!htb]
\begin{center}
\includegraphics[width=3.2in]{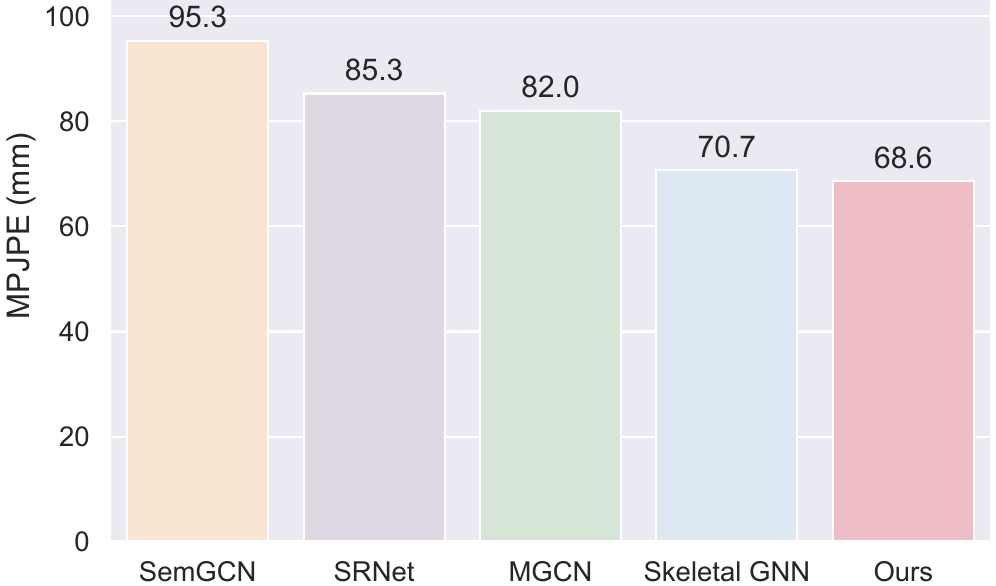}
\end{center}
\caption{Comparison of our model and baselines on the 5\% hardest poses under Protocol \#1.}
\label{Fig:hardPose}
\end{figure}

\medskip\noindent\textbf{Improvements on Hard Poses.}\quad Hard poses, which are characterized by high prediction errors, are specific to the model being used. These poses often have certain inherent characteristics, such as overlapping and self-occlusion. The way in which such cases are dealt with, however, may vary across different models~\cite{Zeng2020SRNet, zeng2021learning, zhao2019semantic}. For instance, when a person is sitting down in a position with their legs crossed, estimating their 3D pose accurately can be difficult due to the complex interactions between different body parts. Our proposed method aims to address this challenge by learning to capture the complex relationships between the joints via the joints mixing MLP layer and GraphWJ mixing layer. As reported in the first table of the main paper, our method yields better performance on hard poses (e.g., Directions, Sitting Down, Photo, and Purchase) compared to recent GCN-based state-of-art methods~\cite{zou2021modulated, zeng2021learning, zhao2019semantic}. In addition, we test our model on the top 5\% hardest poses following ~\cite{Zeng2020SRNet, zeng2021learning}, yielding superior performance over the baselines, as shown in Figure~\ref{Fig:hardPose}.

\end{document}